\documentclass{article}

\usepackage{microtype}
\usepackage{graphicx}
\usepackage{subfigure}
\usepackage{listings}
\usepackage{booktabs} % for professional tables
\usepackage{tabularx}
\usepackage{mathtools} 
\usepackage{amsmath}

\usepackage{hyperref}

\usepackage{cleveref}

\usepackage[accepted]{mlsys2021}

\mlsystitlerunning{Analyzing the Performance of Graph Neural Networks with Pipe Parallelism}

\begin{document}

\twocolumn[
\mlsystitle{Analyzing the Performance of Graph Neural Networks\\ with Pipe Parallelism}

% unless you've provided the [accepted] option to the mlsys2021


\mlsyssetsymbol{equal}{*}

\begin{mlsysauthorlist}
\mlsysauthor{Matthew T. Dearing}{equal,uc}
\mlsysauthor{Xiaoyan (Angela) Wang}{equal,uc}
\end{mlsysauthorlist}

\mlsysaffiliation{uc}{Department of Computer Science, University of Chicago, Chicago, Illinois}

\mlsyscorrespondingauthor{Matthew T. Dearing}{dearing@uchicago.edu}
\mlsyscorrespondingauthor{Xiaoyan (Angela) Wang}{xiaoyanw@uchicago.edu}

\mlsyskeywords{Graph Machine Learning, Graph Neural Networks, Parallelized training, GPipe, PyG, DGL, micro-batching, pipeline parallelism}

\vskip 0.3in

\begin{abstract}
Many interesting datasets ubiquitous in machine learning and deep learning can be described via graphs. As the scale and complexity of graph-structured datasets increase, such as in expansive social networks, protein folding, chemical interaction networks, and material phase transitions, improving the efficiency of the machine learning techniques applied to these is crucial. In this study, we focus on Graph Neural Networks (GNN) that have found great success in tasks such as node or edge classification and link prediction. However, standard GNN models have scaling limits due to necessary recursive calculations performed through dense graph relationships that lead to memory and runtime bottlenecks. While new approaches for processing larger networks are needed to advance graph techniques, and several have been proposed, we study how GNNs could be parallelized using \textit{existing} tools and frameworks that are known to be successful in the deep learning community. In particular, we investigate applying pipeline parallelism to GNN models with \textit{GPipe}, introduced by Google in 2018.
\end{abstract}
]



\printAffiliationsAndNotice{\mlsysEqualContribution} % otherwise use the standard text.

\section{Introduction}
\label{introduction}

Traditional flat or sequential data delivery cannot fully satisfy many of today’s demanding deep learning models, especially as more data structures of interest can be better represented as high-dimensional graphs instead of low-dimensional grids. Graph machine learning has demonstrated successful applications in domains such as chemistry and drug design \cite{duvenaud2015convolutional, Mercado2020}, natural language processing \cite{Vashisht20}, spatio-temporal forecasting \cite{yu2020}, security \cite{zhou2020automating}, social networks \cite{zhou2020automating}, knowledge graphs \cite{arora2020survey}, recommender systems \cite{ying2018}, protein design discovery \cite{strokach2020fast}, and material phase transitions \cite{bapst2020unveiling}. With its increased use, especially on large datasets, performance and scaling challenges with the Graph Neural Network (GNN) \cite{zhou2018graph} are becoming prevalent when using existing machine learning frameworks and accelerators because of memory and data movement limitations \cite{auten2020hardware}

A variety of new solutions to address these issues have been proposed and are highlighted in \Cref{relatedwork}. However, as a practical consideration, leveraging existing state-of-the-art tools and frameworks with demonstrated success at improving deep neural network performance is valuable to push these technologies forward. Therefore, to better understand how training and inference can be more efficient for GNN models, we implement and analyze the performance of parallelized GNN models compared to their unparallelized counterparts when trained on a single CPU and GPU and multiple GPUs. As GNNs are executed sequentially, either layer-by-layer or stage-by-stage, the motivation for this study is to extend current techniques for improving performance by introducing pipeline parallelism into the GNN model architecture \cite{zhang2020architectural}.

\section{Graph Neural Networks} \label{graphneuralnetworks}

The success of deep learning on traditional grid- or sequence-based inputs, such as images and sentences, cannot be overstated. Nevertheless, many datasets in the real-world cannot be expressed within a Euclidean coordinate system, and instead naturally take an arbitrary form of graphs or networks. Various studies exist on how to generalize neural networks for the application to arbitrary irregular graphs \cite{bruna2013spectral, henaff2015deep, duvenaud2015convolutional, li2015gated, defferrard2016convolutional, kipf2016semi}, and we follow the exposition of Kipf and Welling \cite{kipf2016semi} who first introduced the notion of a convolution architecture for a graph.

We consider the general scenario of node classification in a graph. The input is a graph $G = (V, E)$ where $V$ is the set of vertices and $E$ is the set of edges. Each node (or vertex) $i$ in $V$ has a set of features $x_i$. Some pairs of nodes are connected, and the connections are called edges (or links) and form the set $E$. If $n$ is the number of nodes and there are $d$ features in each node, then the set of all features form a $n \times d$ matrix. Some nodes may have labels from a set of $C$ classes. The task is to classify each node into $C$ classes using the ingrained feature information through the edge connectivity between nodes across the graph. 

\begin{table}[htbp]
  \centering
  \caption{Comparisons of training dataset sizes used in this work and considered for future experimentation.}
  \label{table:datasetsizes}
  {\small
\begin{tabular}{|c |ccc|}
    \hline
    \textbf{Dataset} & \textbf{Nodes} & \textbf{Edges} & \textbf{Classes}\\
    %\hhline{~--}
    \hline
    Cora & 2,708 & 5,429 & 7\\
    CiteSeer & 3,312 & 4,732 & 6\\
    PubMed & 19,717 & 44,338 & 3\\
    Reddit & 233,000 & 150,000,000 & 50\\
    Amazon & 6,000,000 & 180,000,000 & 11\\
    \hline
  \end{tabular}
  }
\end{table}

For our analysis, we use the Cora, CiteSeer, and PubMed datasets which are well-established citation network datasets \cite{sen2008collective, yang2016revisiting} often used in benchmark training. For comparison of these in \Cref{table:datasetsizes}, we also list the sizes of two larger datasets, the Reddit post dataset \cite{hamilton2017inductive} and the Amazon data dump \cite{mcauley2015image}. As the data sets used in this study are small and do not require the techniques explored here to provide training efficiency, they offer valuable benchmarks for GNNs and measuring the baseline efficiency of parallelized models.

%\begin{equation}\label{graphlayers}
%\end{equation}

A GNN approaches the node classification problem by building a neural network layer atop a simultaneous message passing paradigm. Suppose there are $L + 1$ layers, $H_0, \ldots, H_{L}$. Then, $H_0 = X$, the input set of features. For each layer $l$, the set of output features $H^{l + 1}$ depends only on the previous layer $H^l$ and the input graph $G$. So, for some efficiently computable function $f$, we have $H^{l + 1} = f(H^l, G)$. Implementations of the GNN model considers different choices of $f$. By setting the output of the last layer to be a single neuron, the model computes the logits for each node, which is then used to classify the nodes. Assuming $f$ is differentiable, this approach can be optimized by standard gradient descent algorithms. In most graph networks studied in the literature, the features $H^{l + 1}(i)$ for node $i$ depend on the original feature $H^l(i)$ and the features of neighboring nodes $H^l(j)$, where $j$ being a neighbor of $i$ connected by edge \{$i$,$j$\}. In some settings, these edges can be of different types, such as being directed or undirected, as well as include features, in which case the edge properties also contribute to the calculation of $f$. The message passing paradigm is designed by how the output features of a layer are \textit{simultaneously updated} based on the input features of the layer, instead of through sequential updates. For many learning tasks on a graph, earlier approaches \cite{dijkstra1959note, cheriton1976finding} usually introduced problem-specific architectures or spectral graph theory to make predictions. However, these algorithms are limited as they require prior knowledge of the graph structure, and the GNN model provides a unified approach that allows for studying the properties of a graph itself.

The experiments presented in this paper are based on the Graph Attention Network (GAT), which is a novel GNN architecture that use attention layers on top of graph convolutions \cite{velivckovic2017graph}. By leveraging this self-attention mechanism, GAT can achieve state-of-the-art accuracy results on several transductive and inductive graph benchmarks including the Cora, CiteSeer, and Pubmed datasets. We chose this model because it can be applied to graphs of different degrees by specifying arbitrary weights on the neighbors, making it useful for a wide variety of graph datasets. GAT is also applicable to inductive learning problems and has generalization capabilities to unseen graph data. Through our experiments, it does not perform as efficiently as simpler Graph Convolutional Networks (GCN), such as those inspired by \cite{kipf2016semi}, lending it to being an illustrative technique to inform us on parallelization benchmarks.

\section{Related Work} \label{relatedwork}

As described above, the core message-passing function of a GNN is the aggregation of features from the neighborhood of each node. Computing a gradient descent operation requires storing the entire graph as a single training batch. With increasing graph size or edge density, the time and memory complexity of this computation can grow exponentially and introduce an information bottleneck \cite{alon2020bottleneck}. This effect limits the scalability of traditional GNNs, many of which, including GAT, do not address these concerns as their benchmark datasets were of a reasonable size for the available device capacities. 

GraphSAGE \cite{hamilton2017inductive} was the first attempt to address graph scalability by using a neighborhood sampling with mini-batch training to limit the number of nodes included in a batch. This approach can introduce redundant calculations as the same nodes may appear in multiple sampled batches and lead to "neighbor explosion" \cite{zeng2020graphsaint}. Similar sampling techniques responded to this challenge by batching sub-graphs instead of nodes, such as in \cite{chiang2019clustergcn}. However, graph clustering approaches are faced with the challenge of defining sub-graphs that sufficiently preserve the edge topology that guides the node feature updates during training, which is an issue directly observed in the analysis of the present study. NeuGraph \cite{ma2019neugraph} introduced parallel computation to enable GNN scalability through a new graph-parallel abstraction of Scatter-ApplyEdge-Gather-ApplyVertex (SAGA-NN). This framework conveniently encapsulates almost all existing GNNs in the literature \cite{zhang2020architectural}, and serves as a foundation for studying parallelized GNN performance optimization. The authors explored the design of a GNN processing framework on top of a dataflow based deep learning system, through which they optimized graph computations, scheduling, and parallelism in a dataflow-based deep learning framework for graphs. Exploring computational efficiencies at the device level, $G^3$ \cite{liu2020g3} is a GNN training framework that implements graph-structured parallel operations that leverage the architectural features on GPUs. By directly utilizing graph-aware components of the GPU, they demonstrated significant speedups in training times over standard implementations in PyTorch and TensorFlow. Finally, a recent approach that avoids graph sampling over nodes or sub-graphs is the Scalable Inception Graph Neural Network (SIGN) \cite{frasca2020sign}. Here, graph convolutional filters of different sizes precompute intermediate node representations. This method enables its scaling to large graphs with classic mini-batching because it retains sufficient expressiveness from the node relationships for effective learning.

\section{Pipeline Parallelism}

Google Brain introduced the scalable pipeline parallelism library \textit{GPipe} \cite{huang2019gpipe} to enable the efficient distributed training of large, memory-consuming deep learning models on current accelerator architectures. According to their published results, \textit{GPipe} increased training times of a 557 million-parameter model by 25 times using eight TPU devices and 3.5 times faster using four devices.

\textit{GPipe} configures a distribution of a sequentially-designed deep neural network across multiple devices. To maximize these devices' capability to calculate in parallel, it further splits the input mini-batch from the training samples into “micro-batches” to distribute across the devices. This micro-batching technique reduces the load for the available memory on the accelerators, resulting in effectively simultaneous training of the same batch across all devices. This approach to pipeline parallelism is like a stacked combination of model parallelism and small data parallelism. During the forward pass, when each partition finishes processing a micro-batch, it shifts the output to the next partition, then immediately begins work on the next micro-batch, enabling partitions can overlap across GPUs. During the backward pass, the gradients for each micro-batch are calculated using the same model parameters from the forward pass, and are consistently accumulated at the end into the mini-batch to update the model parameters. Therefore, the number of partitions separating the data does not affect model quality.

% In summary, \textit{GPipe} splits neural network model layers across multiple accelerators and micro-batches the training batches to increase efficiency in the parallel computations. Hence, \textit{GPipe} can be applied to any deep neural network comprised of sequential layers to perform distributed learning through synchronous stochastic gradient descent and pipeline parallelism. Of particular interest with this approach is the capability to easily scale the performance with additional accelerators without the need to re-tune hyperparameters.

Although designed for deep neural networks, the \textit{GPipe} workflow is applicable to GNNs, with some necessary adaptations that we explore in this study. To the best of our knowledge, this is the first work to consider the idea of applying pipeline parallelism using existing libraries to potentially optimize the runtime of GNNs.

\section{Implementation}

Experiments included training a GAT-based multi-layer sequential neural network on the task of node classification with the citation datasets described above. The PubMed set was solely used to compare performance with pipeline parallelism and graph data batching. The forward propagation model structure remained consistent across all experiments designed with a drop-out layer (\textit{p} = 0.6) followed by a GAT layer with eight heads (attention drop-out = 0.6), a leaky ReLU activation function, a second drop-out layer (\textit{p} = 0.6), a second GAT layer (eight heads, attention drop-out = 0.6) where the outputs are averaged, and, finally, a log softmax function. The neural network model was implemented in PyTorch with the graph frameworks, PyTorch Geometric (PyG) \cite{fey2019fast}, and Deep Graph Library (DGL) \cite{wang2019dgl}. Each framework was compared for performance on each device architecture. Pipeline and data parallelism through \textit{GPipe} was only implemented through DGL. Trials included performance measures for a single CPU, single GPU, and pipe parallel distribution across four GPUs with and without micro-batching of the graph data. For the single device benchmarks, we used an Intel(R) Xeon(R) CPU @ 2.20GHz and NVIDIA Tesla T4 GPU, and four NVIDIA Tesla V100-SXM2 GPUs (DGX) were used for the distributed pipeline parallel experiments.

\textit{GPipe} was incorporated into the GNN models for each framework with the \textit{torchgpipe} \cite{kim2020torchgpipe} library, an implementation of the \textit{GPipe} framework in PyTorch. The defined model is wrapped in a method that takes as parameters a defined distribution of the model layers across the available GPUs as the number of micro-batches (called ``chunks") to be applied. A value of one corresponds to the data parallelism feature being disabled.

% Each layer must take only one argument due to nn.Sequential. There is one more restriction. Every underlying layers’ input and output must be Tensor or Tuple[Tensor, ...]. The reason is that GPipe can’t assume how the non-tensor inputs for a mini-batch can be split for micro-batches. The ability to allow Tuple[Tensor, …] enabled passing the graph through the sequence along with the graph features in a way that GPipe would microbatch.

In Listing~\ref{code:gpipe}, \verb|g| contains the complete graph, \verb|numfeats| represents the number of features per node, and \verb|nclasses| is the number of classes in the classification task. The customizable \verb|balance| array specifies how many layers from the sequence to distribute to each GPU. From this, \textit{GPipe} automatically manages the necessary movements of the model and data across devices. An automated distribution algorithm is also available to optimize this layer assignment. However, for the uniform analysis presented in this paper, we manually set the layer distribution across four devices to ensure consistency for all experiments. With \verb|chunks > 1|, the complete dataset or batches from the training are split into micro-batches by \textit{GPipe} to increase device parallelism. After a partition completes its processing of a micro-batch, it passes the output to the next partition and begins on the next incoming micro-batch in the pipeline. Through this approach, the multiple devices effectively process the same mini-batch (or entire dataset) simultaneously during a training epoch.

\begin{lstlisting}[language=Python, caption={Illustrative \textit{GPipe} implementation with \textit{torchgpipe}.}, label = {code:gpipe}]
import torch.nn as nn
from torchgpipe import GPipe

# Define a sequential model
model = nn.Sequential(
    nn.Dropout(0.6),
    GAT(g, numfeats, 8),
    nn.ELU(),
    nn.Dropout(0.6),
    GAT(g, 8 * 8, nclasses, take_mean = True),
    nn.LogSoftmax(1))

# Wrap the model for pipeline parallelism management
model = GPipe(model, balance = [1, 2, 1, 2], chunks=4)
\end{lstlisting}

A key challenge with this implementation for a GNN is that a sequential module is required for the network layers. A cascade of additional restrictions results, beginning with only a single input of features may be passed through the layers. However, the graph convolution layer expects as input the graph data object and its corresponding features. For our experiments that did not incorporate model parallelism across multiple GPUs, this condition did not pose an issue because we could simply include the full graph data object, \verb|g|, into the GNN model definition and pass the single tensor of features. However, when model parallelism is activated, \textit{GPipe} applies micro-batching to this feature tensor, and the corresponding subset of graph nodes must instead be presented to the graph convolution layer, instead of the full graph data object.

As a workaround for enabling micro-batching, we exploited the option that the sequential module can pass a \textit{single tuple} comprised of multiple tensors. Then, we pass the node indices of the graph as the first tensor along with the corresponding features in a second tensor. \textit{GPipe} applies its micro-batching to each tensor in the tuple, and a subset of graph nodes with the corresponding features are passed along the sequence of layers, as needed. When the graph convolution layer receives the passed tuple, our adapted code extracts the node tensor comprised of the \textit{sub}-graph as determined by the micro-batch from \textit{GPipe}. Both DGL and PyG graph frameworks include a method to re-build a graph structure from a subset of graph nodes, which requires the full graph data object, \verb|g|, for the re-build. The output is then a sub-graph structure expected by the graph convolution layer. The second tensor of the passed tuple that includes the features is subsequently extracted in the graph convolution layer and used in the forward calculation. Upon completion, the two-tensor tuple is reformed with the original nodes of the sub-graph and the updated features to be passed along through the remaining layers of the sequence.

\section{Results}

\begin{table}[h]
\small
\centering
\caption{Benchmark results on multiple compute architectures and graph frameworks for the standard, small graph datasets.}
\label{table:benchmarkA}
\renewcommand{\arraystretch}{1.2}
\begin{tabular}{c|ccc}
\textbf{Compute}  & \multicolumn{3}{c}{\emph{Ave. epoch (ms) \(|\) Test accuracy}} \\
 \textbf{Package} &   \textbf{Cora}    &   \textbf{CiteSeer}   &  \textbf{PubMed}  \\ \hline
CPU -- PyG & 104.4 \(|\) 0.717 & 182.9 \(|\) 0.696 & 798.5 \(|\) 0.718  \\
CPU -- DGL & 71.3 \(|\) 0.785  & 153.4 \(|\) 0.710 & 338.6 \(|\) 0.723  \\
GPU -- PyG & 7.7 \(|\) 0.796 & 8.4 \(|\) 0.720 & 10.9 \(|\) 0.718   \\
GPU -- DGL & 13.3 \(|\) 0.721  & 12.4 \(|\) 0.641 & 12.5 \(|\) 0.682  \\   
\end{tabular}
\end{table}

All experimental training runs were performed for 300 epochs on the same GNN model structure. This model was not optimized for best training performance but remained consistent for all scenarios so that direct comparisons focusing on the graph frameworks, hardware, and parallelism approach could be observed independent of the model structure.

\subsection{Benchmarks}

\begin{table*}[htbp]
        \caption{Benchmarks on different compute architectures and graph frameworks for the sequential GAT model with the PubMed dataset. *The full graph was defined in the GNN model instead of being passed through as a subset of nodes to be re-built as a sub-graph.}
  \label{table:benchmarkB}
  {\small
  \begin{tabular}{|c c|cccccc|}
  \hline
    \textbf{Framework} & \textbf{Compute} & \textbf{Epoch 1 (s)} & \textbf{Epochs 2--300 (s)} & \textbf{Ave. Epoch (s)} & \textbf{Train Loss} & \textbf{Train Acc.} & \textbf{Val Acc.}\\
    \hline
    DGL	& Single CPU & 0.3555 & 101.2 & 0.3386 & 0.2000 & 0.9833 & 0.7520\\
    DGL	& Single GPU & 0.2254 & 3.736 & 0.0125 & 0.2030 & 1.000 & 0.7520\\
    PyG	& Single CPU & 0.7946 & 238.7 & 0.7985 & 0.1567 & 0.9833 & 0.7910\\
    PyG	& Single GPU & 0.2509 & 3.260 & 0.0109 & 0.2131 & 1.000 & 0.7920\\
    DGL	& DGX with GPipe Chunk = 1* & 6.985 & 3.755 & 0.0126 & 0.1984 & 1.000 & 0.7660\\
    PyG	& DGX with GPipe Chunk = 1* & 7.312 & 3.407 & 0.0114 & 0.2097 & 1.000 & 0.7840\\
    DGL	& DGX with GPipe Chunk = 1 & 7.294 & 15.62 & 0.0522 & 0.1879 & 0.9500 & 0.7780\\
    DGL	& DGX with GPipe Chunk = 2 & 7.192 & 12.30 & 0.0411 & 0.4283 & 0.8333 & 0.6000\\
    DGL	& DGX with GPipe Chunk = 3 & 7.281 & 15.29 & 0.0511 & 0.5204 & 0.7667 & 0.4920\\
    DGL	& DGX with GPipe Chunk = 4 & 7.712 & 18.06 & 0.0604 & 0.6016 & 0.7500 & 0.4580\\
    \hline
\end{tabular}
  }
\end{table*}

As a first comparison benchmark, we trained the GNN model on single devices with the citation datasets of Cora, CiteSeer, and PubMed. The training time and test accuracy results are summarized in Table~\ref{table:benchmarkA}. As expected, training times, as measured by the average time per training epoch, on the GPU are faster for both graph frameworks across all datasets. Interestingly, DGL trained on average 35\% faster than PyG on a CPU, while PyG trained on average 29\% faster than DGL on a GPU. This outcome suggests, at least for our applied GNN model training on a single device, PyG may be better optimized for a GPU and DGL for a CPU. Training accuracy remained within a range of 15.5\%, with PyG averaging 2.4\% better than DGL over all datasets.

\begin{figure}[!ht]
\centering
\includegraphics[width=0.95\linewidth]{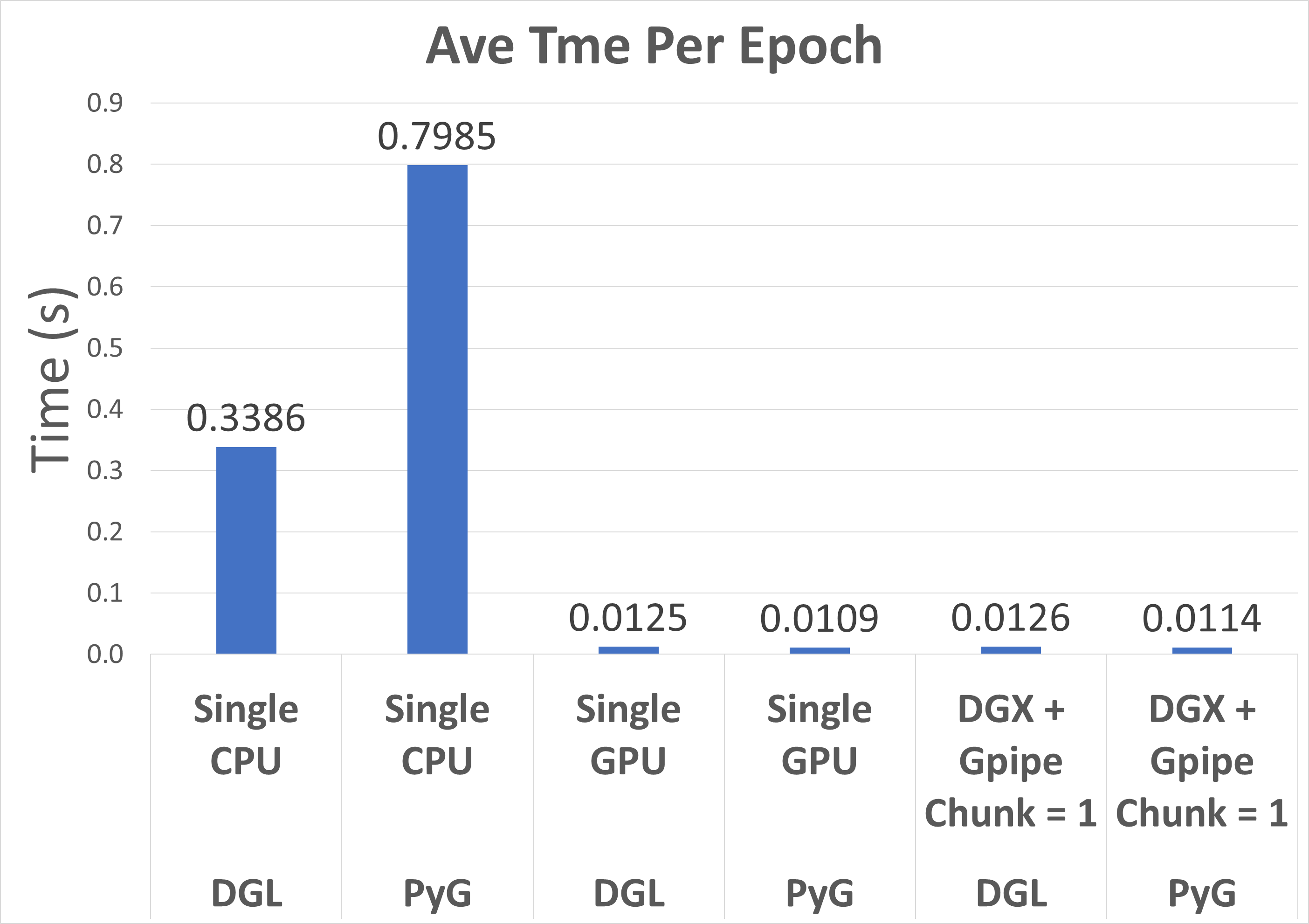}
\caption{Benchmark training times for DGL and PyG on the PubMed dataset comparing the single devices to multiple devices with pipeline parallelism. Here, data parallelism is disabled.}
\label{fig:bechmarksingledevice}
\end{figure}

Next, we compare the average training time per epoch on three compute architectures, including the single CPU and GPU, as previously measured, with the DGX system comprised of four GPUs leveraging \textit{GPipe} pipeline parallelism without micro-batching. In each case presented in ~\Cref{fig:bechmarksingledevice}, the complete graph data object was included in the graph convolution layer during each training epoch. The comprehensive benchmark report in Table~\ref{table:benchmarkB} implements the GAT model on the PubMed dataset across combinations of frameworks and compute clusters for an expanded analysis and comparison of the runtimes and accuracy. We also include the duration of the first epoch in the reported training times to provide a complete comparison of the graph frameworks and hardware, which varied slightly across graph frameworks and architectures. The remaining training epochs ran on the order of 80--100 times faster on the single GPU compared to the single CPU.

\begin{figure}[!ht]
\centering
\includegraphics[width=0.95\linewidth]{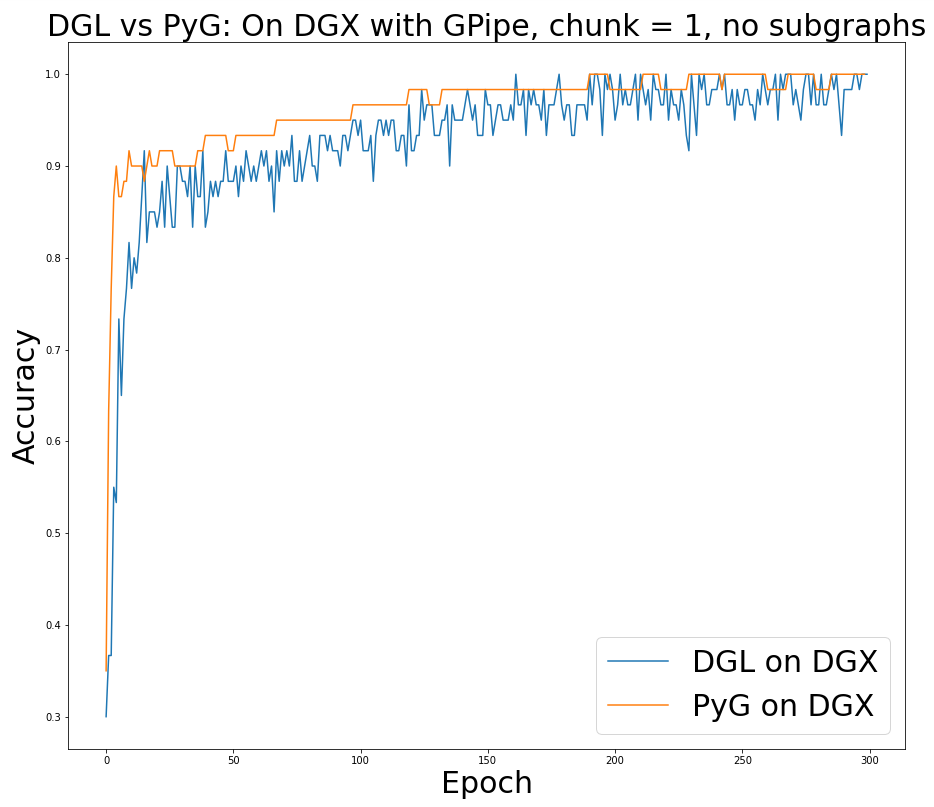}
\caption{Training accuracy with the DGL and PyG frameworks with pipe parallelism across four GPUs with no graph data batching.}
\label{fig:bechmarkaccuracybyframework}
\end{figure}

Surprisingly, no significant performance improvement in training time is observed in the four GPU system using \textit{GPipe} with a “chunk size” = 1 (i.e., no micro-batching) compared to a single GPU. The PubMed dataset used in these experiments is considered small compared to those that are intended to benefit from pipeline parallelism. Therefore, the added cost of shifting data across the four GPUs may overtake the minimal speedup provided by \textit{GPipe}. This may also suggest that the additional feature of data parallelism (via the data “chunks”) provided by \textit{GPipe} is crucial to realizing meaningful performance improvements. We also measured the training accuracy resulting from both graph machine learning frameworks applied with \textit{GPipe} across four GPUs, but without micro-batching, exactly as in the timing measurements. As plotted in Figure~\ref{fig:bechmarkaccuracybyframework}, each framework converged similarly in accuracy over 300 training epochs in this configuration.

\subsection{Increased training time}

\begin{figure}[!ht]
\centering
\includegraphics[width=0.95\linewidth]{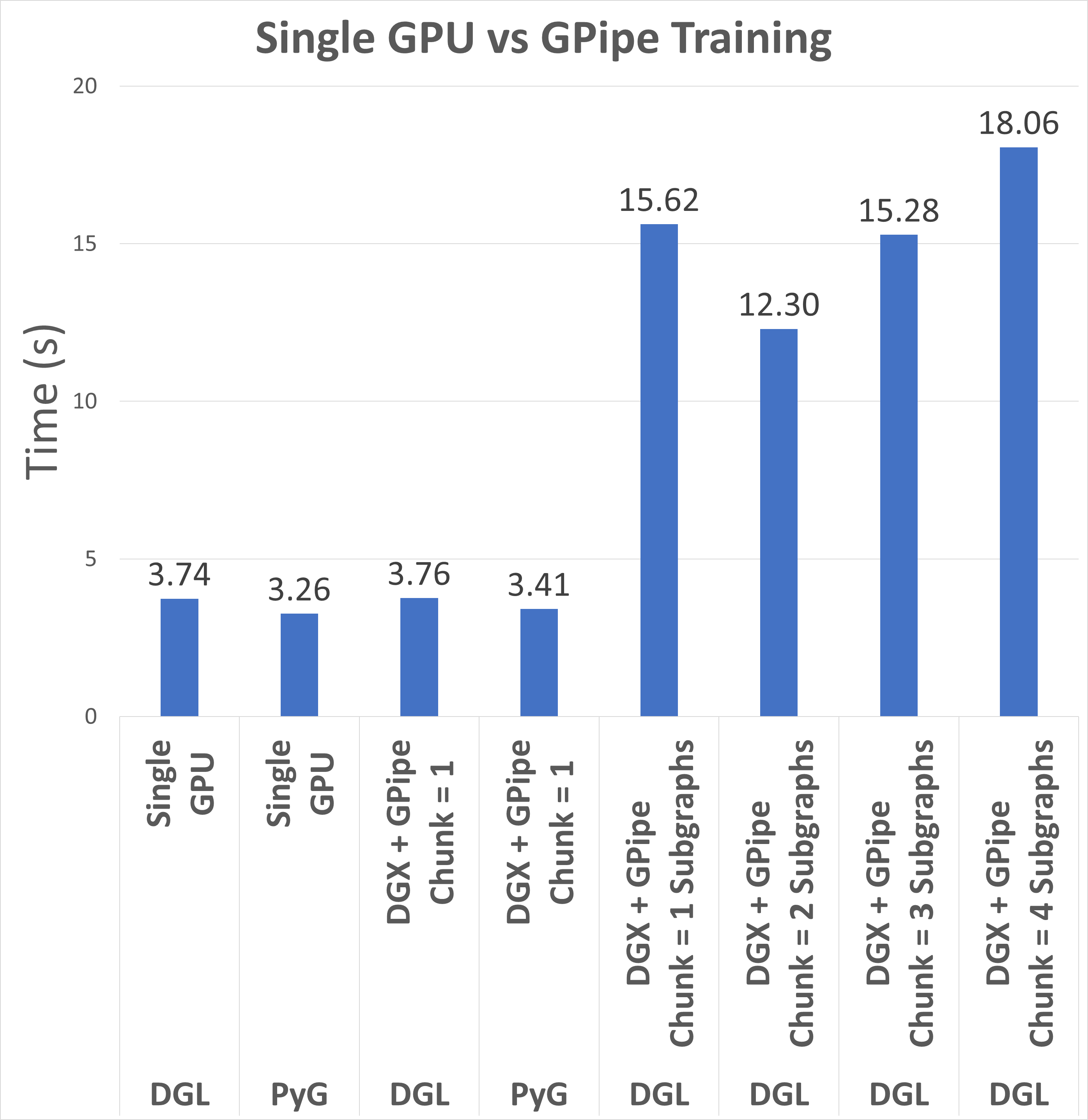}
\caption{Increased training time with \textit{GPipe} applied pipeline parallelism with an increasing number of graph micro-batches.}
\label{fig:bechmarkmultidevice}
\end{figure}

To investigate the impact of data parallelism within \textit{GPipe}, we activated micro-batching and ran the training with the DGL graph framework to compare total training times between a single GPU and multiple distributed GPUs. As seen in ~\Cref{fig:bechmarkmultidevice}, the training times dramatically increase with micro-batching enabled at two, three, and four batches, as generated by \textit{GPipe}.

Our approach adapts the forward training to pass the graph information along with the features through a tuple of tensors into the sequential model. The \textit{GPipe} micro-batching splits each tensor within the tuple so that only a subset of nodes is passed through, along with its corresponding set of features, as expected. The first convolution layer receives this subset of nodes indices but still must have a complete graph structure as its input with the feature tensor. So, a re-build of a graph is first performed with a DGL framework-delivered method. This sub-graph creation from the provided subset of nodes requires the full graph data object as a reference. However, DGL necessitates that the full graph, \verb|g|, must remain on the CPU. To generate the sub-graph within the convolution layer, a copy of the subset node tensor must first be moved from the GPU onto the CPU, then the sub-graph is built and moved back onto the GPU. This data flow across devices was performed twice because our model includes two convolution layers. So, significant overhead was added to the total time just to enable the basic training calculations. As the chunk size increased, more micro-batches were generated, resulting in even more sub-graph build steps. Fortunately, the feature tensor extracted from the passed tuple could remain on the GPU. However, the updated values were still re-packaged into a tuple with the original sub-graph nodes to be returned into the forward pass of the model sequence.

\subsection{Degraded accuracy}

We also observed that the training accuracy suffered severely with an increasing number of micro-batches. Although \textit{GPipe} micro-batching can be disabled, as configured for our benchmark tests (Figures~\ref{fig:bechmarksingledevice} and ~\ref{fig:bechmarkaccuracybyframework}), the expected benefit of pipeline parallelism requires micro-batching. We next ran the same DGL framework-based model with \textit{GPipe} across four GPUs, sequentially distributed as before, to observe the effects of micro-batching in our adapted implementation.

\begin{figure}[!ht]
\centering
\includegraphics[width=0.95\linewidth]{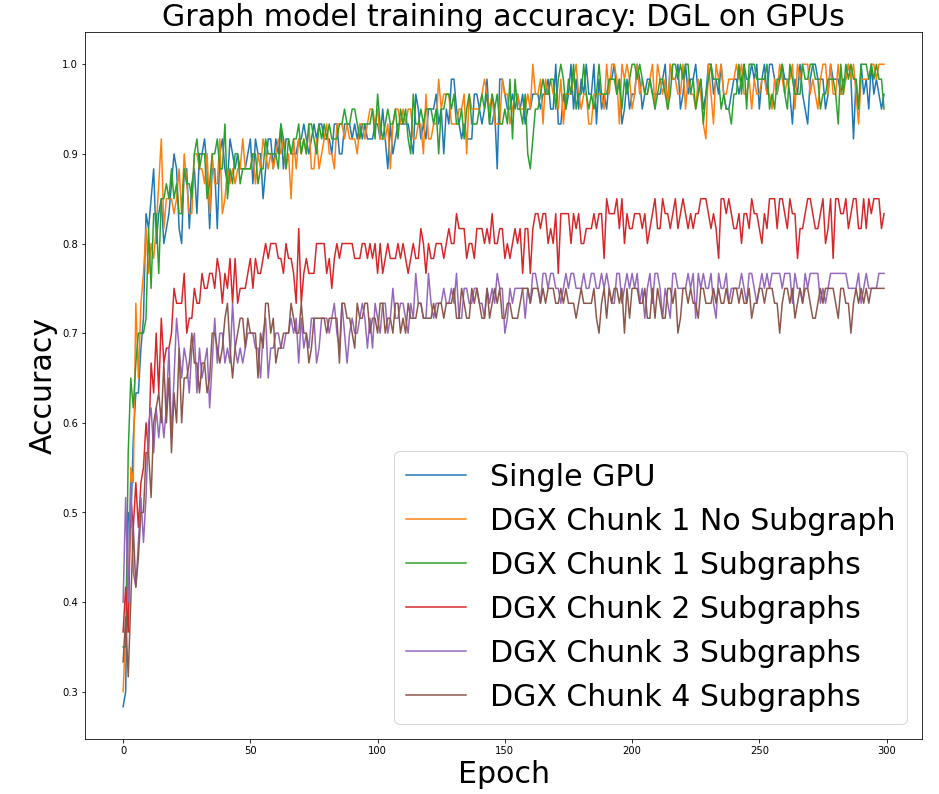}
\caption{Accuracy drop-off with \textit{GPipe} and graph micro-batching with comparisons to the previous training accuracy results without batching.}
\label{fig:degradedaccuracy}
\end{figure}

The intended design of the \textit{GPipe} micro-batching through the \textit{torchgpipe} library implementation is to separate the features tensor into uniform batches. This challenges our adaptation that passes a tuple containing both a node tensor and feature tensor, as we observed the micro-batching being applied to each tensor by \textit{sequentially selecting} the tensor indices into a number of batches equal to the set chunk size parameter. This sequential separation preserves the nodes of the resulting sub-graph with their corresponding features. However, the edge relationships between the nodes are lost. Although edges are re-established during the sub-graph re-build in the convolution layers, the original graph structure is not expected to store its edges sequentially. Therefore, separating the graph this way during the micro-batching likely eliminates crucial node relationships that need aggregated during the graph convolution layer calculations. As expected from such a potential for significant information loss during the \textit{GPipe} micro-batching, as the number of batches generated increases, the training accuracy drops, as seen in ~\Cref{fig:degradedaccuracy}.

% On the other hand, both DGL and PyG provide more sophisticated approaches to generate mini-batches from the complete graph structures. For example, available in DGL as of v0.5.3, stochastic mini-batch training can be applied when not all the features from all nodes must be stored on a single GPU. This approach takes a sample of the neighborhood about a node from its previous layer and iterates until the sample search reaches the input. From this process, a dependency graph is built backward from the output to the input, maintaining the relationship structure throughout the training for a single node.

% The initial intention of our implementation was to leverage these stochastic mini-batch “blocks,” as they are referred to in DGL, as the batches for \textit{GPipe}. However, \textit{GPipe} sequentially split these blocks into smaller batches, as they did on the sub-set of nodes tensor, so the relationships from the predetermined dependency graph is still lost.

% In addition, DGL also offers a distributed training mechanism that uses its stochastic mini-batch blocks to parallelize the computation and partition graph data during training. However, at the time of this study, the distributed training in DGL only supports homogeneous graph structures with one node type and one edge type. So, we could not compare this technique with our benchmark datasets.

\section{Conclusion and Future Work}

By analyzing the performance of GNNs using pipeline parallelism via \textit{GPipe}, our first results suggest that although \textit{GPipe} has demonstrated great success in optimizing deep neural networks, its ability to deliver efficiency for a graph neural network remains limited with the adapted implementation presented here.

An immediate scope for future work is to determine how to customize the \textit{GPipe} data parallelism to utilize intelligent graph batching instead of a sequential separation by index. Such an improvement is expected to increase accuracy to match the benchmark levels while benefiting from the parallelism in runtime efficiency. The SIGN technique described in \Cref{relatedwork} may be the best batching approach to consider for parallelizing GNNs with our implementation because it avoids the experienced pitfalls of node and graph sampling and instead provides precomputed node representations that may be straightforwardly mini-batched by \textit{GPipe}.

Pipeline parallelism is intended to benefit neural network training on very large datasets, much greater in size than the PubMed set used here to establish our adapted implementation without overburdening memory resources. We anticipate that runtime performance will increase for training on extremely large graphs after memory bottlenecks or computational complexity overtake the capability offered by a single GPU. Extending the current implementation to massive datasets, on the scale of millions of nodes and a billion edges, such as the Reddit post dataset \cite{hamilton2017inductive}, the Amazon data dump \cite{mcauley2015image}, and others available with data loaders for DGL and PyG through the Open Graph Benchmark (OGB) \cite{hu2020ogb}, will better illustrate the impact of \textit{GPipe} parallelism on GNNs, and provide a deeper understanding for potential enhancements to our current implementation for improving training performance with these existing tools and frameworks.

% \section*{Acknowledgements}
\section*{Acknowledgments}
Thank you to Prof. Ian Foster, Prof. Rick Stevens, and Peng Ding for the guidance and feedback on this paper. We also thank the DGX team, Daniel Murphy-Olson and Ryan Aydelott, and the Computing, Environment, and Life Sciences directorate at Argonne National Laboratory.

% \textbf{Do not} include acknowledgements in the initial version of
% the paper submitted for blind review.

% If a paper is accepted, the final camera-ready version can (and
% probably should) include acknowledgements. In this case, please
% place such acknowledgements in an unnumbered section at the
% end of the paper. Typically, this will include thanks to reviewers
% who gave useful comments, to colleagues who contributed to the ideas,
% and to funding agencies and corporate sponsors that provided financial

\nocite{langley00}

\bibliographystyle{mlsys2021}
\bibliography{ref}

\begin{thebibliography}{35}
\providecommand{\natexlab}[1]{#1}
\providecommand{\url}[1]{\texttt{#1}}
\expandafter\ifx\csname urlstyle\endcsname\relax
  \providecommand{\doi}[1]{doi: #1}\else
  \providecommand{\doi}{doi: \begingroup \urlstyle{rm}\Url}\fi

\bibitem[Alon \& Yahav(2020)Alon and Yahav]{alon2020bottleneck}
Alon, U. and Yahav, E.
\newblock On the bottleneck of graph neural networks and its practical
  implications.
\newblock \emph{arXiv preprint arXiv:2006.05205}, 2020.

\bibitem[Arora(2020)]{arora2020survey}
Arora, S.
\newblock A survey on graph neural networks for knowledge graph completion,
  2020.

\bibitem[Auten et~al.(2020)Auten, Tomei, and Kumar]{auten2020hardware}
Auten, A., Tomei, M., and Kumar, R.
\newblock Hardware acceleration of graph neural networks.
\newblock In \emph{2020 57th ACM/IEEE Design Automation Conference (DAC)}, pp.\
   1--6. IEEE, 2020.

\bibitem[Bapst et~al.(2020)Bapst, Keck, Grabska-Barwińska, Donner, Cubuk,
  Schoenholz, Obika, Nelson, Back, Hassabis, and Kohli]{bapst2020unveiling}
Bapst, V., Keck, T., Grabska-Barwińska, A., Donner, C., Cubuk, E., Schoenholz,
  S., Obika, A., Nelson, A., Back, T., Hassabis, D., and Kohli, P.
\newblock Unveiling the predictive power of static structure in glassy systems.
\newblock \emph{Nature Physics}, 16\penalty0 (4):\penalty0 448--454, 2020.

\bibitem[Bruna et~al.(2013)Bruna, Zaremba, Szlam, and LeCun]{bruna2013spectral}
Bruna, J., Zaremba, W., Szlam, A., and LeCun, Y.
\newblock Spectral networks and locally connected networks on graphs.
\newblock \emph{arXiv preprint arXiv:1312.6203}, 2013.

\bibitem[Cheriton \& Tarjan(1976)Cheriton and Tarjan]{cheriton1976finding}
Cheriton, D. and Tarjan, R.
\newblock Finding minimum spanning trees.
\newblock \emph{SIAM Journal on Computing}, 5\penalty0 (4):\penalty0 724--742,
  1976.

\bibitem[Chiang et~al.(2019)Chiang, Liu, Si, Li, Bengio, and
  Hsieh]{chiang2019clustergcn}
Chiang, W.-L., Liu, X., Si, S., Li, Y., Bengio, S., and Hsieh, C.-J.
\newblock Cluster-gcn: An efficient algorithm for training deep and large graph
  convolutional networks.
\newblock In \emph{Proceedings of the 25th ACM SIGKDD International Conference
  on Knowledge Discovery and Data Mining}, pp.\  257--266, 2019.

\bibitem[Defferrard et~al.(2016)Defferrard, Bresson, and
  Vandergheynst]{defferrard2016convolutional}
Defferrard, M., Bresson, X., and Vandergheynst, P.
\newblock Convolutional neural networks on graphs with fast localized spectral
  filtering.
\newblock In \emph{Advances in neural information processing systems}, pp.\
  3844--3852, 2016.

\bibitem[Dijkstra(1959)]{dijkstra1959note}
Dijkstra, E.
\newblock A note on two problems in connexion with graphs.
\newblock \emph{Numerische Mathematik}, 1:\penalty0 269--271, 1959.

\bibitem[Duvenaud et~al.(2015)Duvenaud, Maclaurin, Iparraguirre, Bombarell,
  Hirzel, Aspuru-Guzik, and Adams]{duvenaud2015convolutional}
Duvenaud, D.~K., Maclaurin, D., Iparraguirre, J., Bombarell, R., Hirzel, T.,
  Aspuru-Guzik, A., and Adams, R.~P.
\newblock Convolutional networks on graphs for learning molecular fingerprints.
\newblock In \emph{Advances in neural information processing systems}, pp.\
  2224--2232, 2015.

\bibitem[Fey \& Lenssen(2019)Fey and Lenssen]{fey2019fast}
Fey, M. and Lenssen, J.~E.
\newblock Fast graph representation learning with pytorch geometric.
\newblock \emph{arXiv preprint arXiv:1903.02428}, 2019.

\bibitem[Frasca et~al.(2020)Frasca, Rossi, Eynard, Chamberlain, Bronstein, and
  Monti]{frasca2020sign}
Frasca, F., Rossi, E., Eynard, D., Chamberlain, B., Bronstein, M., and Monti,
  F.
\newblock Sign: Scalable inception graph neural network.
\newblock \emph{arXiv preprint arXiv:2004.11198}, 2020.

\bibitem[Hamilton et~al.(2017)Hamilton, Ying, and
  Leskovec]{hamilton2017inductive}
Hamilton, W., Ying, Z., and Leskovec, J.
\newblock Inductive representation learning on large graphs.
\newblock In \emph{Advances in neural information processing systems}, pp.\
  1024--1034, 2017.

\bibitem[Henaff et~al.(2015)Henaff, Bruna, and LeCun]{henaff2015deep}
Henaff, M., Bruna, J., and LeCun, Y.
\newblock Deep convolutional networks on graph-structured data.
\newblock \emph{arXiv preprint arXiv:1506.05163}, 2015.

\bibitem[Hu et~al.(2020)Hu, Fey, Zitnik, Dong, Ren, Liu, Catasta, and
  Leskovec]{hu2020ogb}
Hu, W., Fey, M., Zitnik, M., Dong, Y., Ren, H., Liu, B., Catasta, M., and
  Leskovec, J.
\newblock Open graph benchmark: Datasets for machine learning on graphs.
\newblock \emph{arXiv preprint arXiv:2005.00687}, 2020.

\bibitem[Huang et~al.(2019)Huang, Cheng, Bapna, Firat, Chen, Chen, Lee, Ngiam,
  Le, Wu, et~al.]{huang2019gpipe}
Huang, Y., Cheng, Y., Bapna, A., Firat, O., Chen, D., Chen, M., Lee, H., Ngiam,
  J., Le, Q.~V., Wu, Y., et~al.
\newblock Gpipe: Efficient training of giant neural networks using pipeline
  parallelism.
\newblock In \emph{Advances in neural information processing systems}, pp.\
  103--112, 2019.

\bibitem[Kim et~al.(2020)Kim, Lee, Jeong, Baek, Yoon, Kim, Lim, and
  Kim]{kim2020torchgpipe}
Kim, C., Lee, H., Jeong, M., Baek, W., Yoon, B., Kim, I., Lim, S., and Kim, S.
\newblock torchgpipe: On-the-fly pipeline parallelism for training giant
  models.
\newblock 2020.

\bibitem[Kipf \& Welling(2016)Kipf and Welling]{kipf2016semi}
Kipf, T.~N. and Welling, M.
\newblock Semi-supervised classification with graph convolutional networks.
\newblock \emph{arXiv preprint arXiv:1609.02907}, 2016.

\bibitem[Li et~al.(2015)Li, Tarlow, Brockschmidt, and Zemel]{li2015gated}
Li, Y., Tarlow, D., Brockschmidt, M., and Zemel, R.
\newblock Gated graph sequence neural networks.
\newblock \emph{arXiv preprint arXiv:1511.05493}, 2015.

\bibitem[Liu et~al.(2020)Liu, Lu, Chen, and He]{liu2020g3}
Liu, H., Lu, S., Chen, X., and He, B.
\newblock G3: when graph neural networks meet parallel graph processing systems
  on gpus.
\newblock \emph{Proceedings of the VLDB Endowment}, 13\penalty0 (12):\penalty0
  2813--2816, 2020.

\bibitem[Ma et~al.(2019)Ma, Yang, Miao, Xue, Wu, Zhou, and Dai]{ma2019neugraph}
Ma, L., Yang, Z., Miao, Y., Xue, J., Wu, M., Zhou, L., and Dai, Y.
\newblock Neugraph: parallel deep neural network computation on large graphs.
\newblock In \emph{2019 $\{$USENIX$\}$ Annual Technical Conference
  ($\{$USENIX$\}$$\{$ATC$\}$ 19)}, pp.\  443--458, 2019.

\bibitem[McAuley et~al.(2015)McAuley, Targett, Shi, and Van
  Den~Hengel]{mcauley2015image}
McAuley, J., Targett, C., Shi, Q., and Van Den~Hengel, A.
\newblock Image-based recommendations on styles and substitutes.
\newblock In \emph{Proceedings of the 38th international ACM SIGIR conference
  on research and development in information retrieval}, pp.\  43--52, 2015.

\bibitem[Mercado et~al.(2020)Mercado, Rastemo, Lindelöf, Klambauer, Engkvist,
  Chen, and Bjerrum]{Mercado2020}
Mercado, R., Rastemo, T., Lindelöf, E., Klambauer, G., Engkvist, O., Chen, H.,
  and Bjerrum, E.~J.
\newblock {Graph Networks for Molecular Design}.
\newblock 8 2020.
\newblock \doi{10.26434/chemrxiv.12843137.v1}.
\newblock URL
  \url{https://chemrxiv.org/articles/preprint/Graph_Networks_for_Molecular_Design/12843137}.

\bibitem[Sen et~al.(2008)Sen, Namata, Bilgic, Getoor, Galligher, and
  Eliassi-Rad]{sen2008collective}
Sen, P., Namata, G., Bilgic, M., Getoor, L., Galligher, B., and Eliassi-Rad, T.
\newblock Collective classification in network data.
\newblock \emph{AI magazine}, 29\penalty0 (3):\penalty0 93--93, 2008.

\bibitem[Strokach et~al.(2020)Strokach, Becerra, Corbi-Verge, Perez-Riba, and
  Kim]{strokach2020fast}
Strokach, A., Becerra, D., Corbi-Verge, C., Perez-Riba, A., and Kim, P.
\newblock Fast and flexible protein design using deep graph neural network.
\newblock \emph{Cell Systems}, 11\penalty0 (4):\penalty0 402--411.e4, 2020.

\bibitem[Vashishth(2019)]{Vashisht20}
Vashishth, S.
\newblock Neural graph embedding methods for natural language processing.
\newblock \emph{CoRR}, abs/1911.03042, 2019.
\newblock URL \url{http://arxiv.org/abs/1911.03042}.

\bibitem[Veli{\v{c}}kovi{\'c} et~al.(2017)Veli{\v{c}}kovi{\'c}, Cucurull,
  Casanova, Romero, Lio, and Bengio]{velivckovic2017graph}
Veli{\v{c}}kovi{\'c}, P., Cucurull, G., Casanova, A., Romero, A., Lio, P., and
  Bengio, Y.
\newblock Graph attention networks.
\newblock \emph{arXiv preprint arXiv:1710.10903}, 2017.

\bibitem[Wang et~al.(2019)Wang, Zheng, Ye, Gan, Li, Song, Zhou, Ma, Yu, Gai,
  Xiao, He, Karypis, Li, and Zhang]{wang2019dgl}
Wang, M., Zheng, D., Ye, Z., Gan, Q., Li, M., Song, X., Zhou, J., Ma, C., Yu,
  L., Gai, Y., Xiao, T., He, T., Karypis, G., Li, J., and Zhang, Z.
\newblock Deep graph library: A graph-centric, highly-performant package for
  graph neural networks.
\newblock \emph{arXiv preprint arXiv:1909.01315}, 2019.

\bibitem[Yang et~al.(2016)Yang, Cohen, and Salakhudinov]{yang2016revisiting}
Yang, Z., Cohen, W., and Salakhudinov, R.
\newblock Revisiting semi-supervised learning with graph embeddings.
\newblock In \emph{International conference on machine learning}, pp.\  40--48.
  PMLR, 2016.

\bibitem[Ying et~al.(2018)Ying, He, Chen, Eksombatchai, Hamilton, and
  Leskovec]{ying2018}
Ying, R., He, R., Chen, K., Eksombatchai, P., Hamilton, W.~L., and Leskovec, J.
\newblock Graph convolutional neural networks for web-scale recommender
  systems.
\newblock \emph{CoRR}, abs/1806.01973, 2018.
\newblock URL \url{http://arxiv.org/abs/1806.01973}.

\bibitem[Yu et~al.(2017)Yu, Yin, and Zhu]{yu2020}
Yu, B., Yin, H., and Zhu, Z.
\newblock Spatio-temporal graph convolutional neural network: {A} deep learning
  framework for traffic forecasting.
\newblock \emph{CoRR}, abs/1709.04875, 2017.
\newblock URL \url{http://arxiv.org/abs/1709.04875}.

\bibitem[Zeng et~al.(2020)Zeng, Zhou, Srivastava, Kannan, and
  Prasanna]{zeng2020graphsaint}
Zeng, H., Zhou, H., Srivastava, A., Kannan, R., and Prasanna, V.
\newblock Graphsaint: Graph sampling based inductive learning method.
\newblock \emph{arXiv preprint arXiv:1907.04931}, 2020.

\bibitem[Zhang et~al.(2020)Zhang, Leng, Ma, Miao, Li, and
  Guo]{zhang2020architectural}
Zhang, Z., Leng, J., Ma, L., Miao, Y., Li, C., and Guo, M.
\newblock Architectural implications of graph neural networks.
\newblock \emph{IEEE Computer Architecture Letters}, 19\penalty0 (1):\penalty0
  59--62, 2020.

\bibitem[Zhou et~al.(2018)Zhou, Cui, Zhang, Yang, Liu, Wang, Li, and
  Sun]{zhou2018graph}
Zhou, J., Cui, G., Zhang, Z., Yang, C., Liu, Z., Wang, L., Li, C., and Sun, M.
\newblock Graph neural networks: A review of methods and applications.
\newblock \emph{arXiv preprint arXiv:1812.08434}, 2018.

\bibitem[Zhou et~al.(2020)Zhou, Xu, Rush, and Yu]{zhou2020automating}
Zhou, J., Xu, Z., Rush, A.~M., and Yu, M.
\newblock Automating botnet detection with graph neural networks.
\newblock \emph{arXiv preprint arXiv:2003.06344}, 2020.

\end{thebibliography}

%%%%%%%%%%%%%%%%%%%%%%%%%%%%%%%%%%%%%%%%%%%%%%%%%%%%%%%%%%%%%%%%%%%%%%%%%%%%%%%
%%%%%%%%%%%%%%%%%%%%%%%%%%%%%%%%%%%%%%%%%%%%%%%%%%%%%%%%%%%%%%%%%%%%%%%%%%%%%%%
% SUPPLEMENTAL CONTENT AS APPENDIX AFTER REFERENCES
%%%%%%%%%%%%%%%%%%%%%%%%%%%%%%%%%%%%%%%%%%%%%%%%%%%%%%%%%%%%%%%%%%%%%%%%%%%%%%%
%%%%%%%%%%%%%%%%%%%%%%%%%%%%%%%%%%%%%%%%%%%%%%%%%%%%%%%%%%%%%%%%%%%%%%%%%%%%%%%

%%%%%%%%%%%%%%%%%%%%%%%%%%%%%%%%%%%%%%%%%%%%%%%%%%%%%%%%%%%%%%%%%%%%%%%%%%%%%%%
%%%%%%%%%%%%%%%%%%%%%%%%%%%%%%%%%%%%%%%%%%%%%%%%%%%%%%%%%%%%%%%%%%%%%%%%%%%%%%%

\end{document}